\documentclass{ieeeaccess}
\usepackage{cite}
\usepackage{amsmath,amssymb,amsfonts}
\usepackage{algorithmic}
\usepackage{graphicx}
\usepackage{textcomp}
\usepackage{CJKutf8}
\usepackage{makecell}
\usepackage{booktabs}
\usepackage{arydshln}
\usepackage{eqnarray}
\usepackage{color}
\usepackage{mathrsfs}
\usepackage{multirow}
\usepackage{graphicx}

\usepackage{bm}
\makeatletter
\AtBeginDocument{\DeclareMathVersion{bold}
\SetSymbolFont{operators}{bold}{T1}{times}{b}{n}
\SetSymbolFont{NewLetters}{bold}{T1}{times}{b}{it}
\SetMathAlphabet{\mathrm}{bold}{T1}{times}{b}{n}
\SetMathAlphabet{\mathit}{bold}{T1}{times}{b}{it}
\SetMathAlphabet{\mathbf}{bold}{T1}{times}{b}{n}
\SetMathAlphabet{\mathtt}{bold}{OT1}{pcr}{b}{n}
\SetSymbolFont{symbols}{bold}{OMS}{cmsy}{b}{n}
\renewcommand\boldmath{\@nomath\boldmath\mathversion{bold}}}
\makeatother

\def\BibTeX{{\rm B\kern-.05em{\sc i\kern-.025em b}\kern-.08em
    T\kern-.1667em\lower.7ex\hbox{E}\kern-.125emX}}

\begin{document}
\history{Date of publication xxxx 00, 0000, date of current version xxxx 00, 0000.}
\doi{10.1109/ACCESS.2024.0429000}

\title{IAE: Irony-based Adversarial Examples for Sentiment Analysis Systems}
\author{\uppercase{Xiaoyin Yi}\authorrefmark{1,2},and
\uppercase{Jiacheng Huang}\authorrefmark{2}}

\address[1]{Chongqing Key Laboratory of Public Big Data Security Technology,Chongqing,400000, China}
\address[2]{School of Computer Science and Technology, Chongqing University of Posts and Telecommunications,Chongqing, 400065, China}

\tfootnote{This research was funded by the Scientific and Technological Research Program of Chongqing Municipal Education Commission (Grant No. KJQN202302403).}

\markboth
{Author \headeretal: Preparation of Papers for IEEE TRANSACTIONS and JOURNALS}
{Author \headeretal: Preparation of Papers for IEEE TRANSACTIONS and JOURNALS}

\corresp{Corresponding author: Jiacheng Huang (e-mail: Dylan.JiaCheng.Huang@outlook.com).}

\begin{abstract}
Adversarial examples, which are inputs deliberately perturbed with imperceptible changes to induce model errors, have raised serious concerns for the reliability and security of deep neural networks (DNNs). 
While adversarial attacks have been extensively studied in continuous data domains such as images, the discrete nature of text presents unique challenges. 
In this paper, we propose Irony-based Adversarial Examples (IAE), a method that transforms straightforward sentences into ironic ones to create adversarial text. 
This approach exploits the rhetorical device of irony, where the intended meaning is opposite to the literal interpretation, requiring a deeper understanding of context to detect. 
The IAE method is particularly challenging due to the need to accurately locate evaluation words, substitute them with appropriate collocations, and expand the text with suitable ironic elements while maintaining semantic coherence. 
Our research makes the following key contributions: 
(1) We introduce IAE, a strategy for generating textual adversarial examples using irony. 
This method does not rely on pre-existing irony corpora, making it a versatile tool for creating adversarial text in various NLP tasks. 
(2) We demonstrate that the performance of several state-of-the-art deep learning models on sentiment analysis tasks significantly deteriorates when subjected to IAE attacks. This finding underscores the susceptibility of current NLP systems to adversarial manipulation through irony. 
(3) We compare the impact of IAE on human judgment versus NLP systems, revealing that humans are less susceptible to the effects of irony in text. 
\end{abstract}


\titlepgskip=-21pt

\maketitle

\section{Introduction}
Adversarial examples~\cite{Goodfellow14}, crafted by adding imperceptible tiny perturbations 
to origin inputs maliciously, cause deep neural networks (DNNs) to fail blatantly. 
The secure issue, namely adversarial attack, is being widely concerned among researchers as soon as it was proposed. 
Extensive research has revealed that adversarial examples widely exist in many fields, 
e.g., computer vision (CV)~\cite{Su19}, natural language processing (NLP)~\cite{Jia17} and automatic speech recognition (ASR)~\cite{Wang21}.

Textual data is not as continuous as images which are capable of being perturbed imperceptibly with pixel noise. 
Instead, it is impossible to craft a factual imperceptible perturbation on a text due to its discrete nature.
Furthermore, the grammar and semantics may be broken easily by changing even a character. 
The textual adversarial attack is confronted with greater challenges compared with images.

A variety of textual adversarial attack models has been proposed in many NLP tasks, 
incorporating machine translation~\cite{Belinkov18}, question-answering system~\cite{Jia17}, sentiment analysis~\cite{Jin20}, et al. 
Spelling mistake~\cite{Ebrahimi18}, visually similar characters substitution~\cite{Eger19}, synonyms substitution~\cite{Ren19} and sentence paraphrasing~\cite{Iyyer18} 
are typical textual adversarial attack methods ranging from word-level to sentence-level while categorized by 
attacking granularity. However, there are still a few issues while assuming those methods in practical situations: 
1) Subtle spelling mistakes can be recovered easily with spelling error correction~\cite{Pruthi19}. 
2) Words out of vocabulary may arise attention and alertness while exceeding averages in a text. 
3) Word substitution and sentence paraphrasing may cause grammar to be broken or semantics deviated. 
Therefore, we consider a textual adversarial attacking method more practically.

\begin{table}[htbp]
    \setlength{\tabcolsep}{3pt}
	\centering
	\caption{\label{table:examples} Examples of straightforward and ironic text.}
	\begin{tabular}{p{8.0cm}}
	  \toprule
		Straightforward 1: \begin{CJK}{UTF8}{gbsn}他真是个\textcolor{blue}{糟糕}的守门员，让对方进了六个球。\end{CJK} 
		He is a really \textcolor{blue}{terrible} goalkeeper, allowing the other side to score six goals.\\
	  \specialrule{0em}{1pt}{1pt}
	  \hdashline[3pt/3pt]
	  \specialrule{0em}{1pt}{1pt}
		Ironic 1: \begin{CJK}{UTF8}{gbsn}他真是个\textcolor{red}{有天赋}的守门员，让对方进了六个球。\end{CJK} 
		He is a really \textcolor{red}{talented} goalkeeper, allowing the other side to score six goals.\\
	  \hline
	  \specialrule{0em}{1pt}{1pt}
		Straightforward 2: \begin{CJK}{UTF8}{gbsn}那个男人真\textcolor{blue}{恶心}，在公共场所随地吐痰。\end{CJK} 
		That man is totally \textcolor{blue}{disgusting}, spiting everywhere in public.\\
	  \specialrule{0em}{1pt}{1pt}
	  \hdashline[3pt/3pt]
	  \specialrule{0em}{1pt}{1pt}
		Ironic 2(a): \begin{CJK}{UTF8}{gbsn}那个男人真\textcolor{red}{美味}，在共场所随地吐痰。\end{CJK} 
		That man is totally \textcolor{red}{delicious}, spiting everywhere in public. \\
	  \specialrule{0em}{1pt}{1pt}
	  \hdashline[3pt/3pt]
	  \specialrule{0em}{1pt}{1pt}
		Ironic 2(b): \begin{CJK}{UTF8}{gbsn}那个男人真\textcolor{red}{优雅}，在共场所随地吐痰。\end{CJK} 
		That man is totally \textcolor{red}{elegant}, spiting everywhere in public.\\
	  \specialrule{0em}{1pt}{1pt}
	  \hdashline[3pt/3pt]
	  \specialrule{0em}{1pt}{1pt}
		Ironic 2(c): \begin{CJK}{UTF8}{gbsn}那个男人真\textcolor{red}{优雅}，在共场所随地吐痰。\textcolor{red}{真是值得称赞啊。}\end{CJK} 
		That man is totally \textcolor{red}{elegant}, spiting everywhere in public.\textcolor{red}{\ It is really praiseworthy.}\\
	  \bottomrule
	\end{tabular}
  \end{table}

  The irony is a kind of rhetorical device expressing a strong emotion referring to the opposite of literal meaning and needs to 
  understand the actual meaning from context. Detecting irony is challenging while implementing it the model needs to have 
  human-level language understanding ability. As far as we know, there are no studies considering converting text from straightforward to ironic as a method 
  of generating textual adversarial examples orienting the NLP task of sentiment analysis presently. 
  
  The cruxes of converting a text from straightforward into ironic are to turn the polarity of the evaluation words and 
  make an ironic expansion appropriately when necessary. Specifically, there are at least three challenges here: 
  1) locating evaluation words, 
  2) substituting evaluation words with correct collocation, and 
  3) expanding text with appropriate ironic evaluation.
  
  Without loss of generality, we consider Chinese irony-based adversarial examples in this paper. As shown in Table~\ref{table:examples}, 
  Chinese words \begin{CJK}{UTF8}{gbsn}``糟糕''\end{CJK} is an evaluation to \begin{CJK}{UTF8}{gbsn}``守门员''\end{CJK} in first sentence, 
  where \begin{CJK}{UTF8}{gbsn}``糟糕''\end{CJK} means ``\emph{terrible}'' and \begin{CJK}{UTF8}{gbsn}``守门员''\end{CJK} means ``\emph{goalkeeper}''. 
  It is necessary to locate the words \begin{CJK}{UTF8}{gbsn}``糟糕''\end{CJK} as an evaluation disclosing negative emotion and 
  then substitute \begin{CJK}{UTF8}{gbsn}``糟糕''\end{CJK} with \begin{CJK}{UTF8}{gbsn}``有天赋''\end{CJK} which means ``\emph{talented}''. 
  Humans are in capable of understanding the second sentence still exhibiting negative emotion with strong language 
  comprehending ability, although the evaluation words \begin{CJK}{UTF8}{gbsn}``有天赋''\end{CJK} is an absolutely positive evaluation 
  literally. Besides, it ought to be notice the substitution needs to consider collocation relation instead of substituting with 
  antonym simply. For example, \begin{CJK}{UTF8}{gbsn}``美味''\end{CJK} is one of antonyms for \begin{CJK}{UTF8}{gbsn}``恶心''\end{CJK}, 
  where \begin{CJK}{UTF8}{gbsn}``美味''\end{CJK} means ``\emph{delicious}'' and \begin{CJK}{UTF8}{gbsn}``恶心''\end{CJK} means ``\emph{disgusting}'', 
  but \begin{CJK}{UTF8}{gbsn}``美味''\end{CJK} is not supposed to collocate with \begin{CJK}{UTF8}{gbsn}``男人''\end{CJK}, which means ``\emph{man}'', 
  referring to the context in fourth sentence, and it is supposed to be substituted with \begin{CJK}{UTF8}{gbsn}``优雅''\end{CJK} instead, 
  which means ``\emph{elegant}'', as shown in fifth sentence. Furthermore, the whole sentence needs to be semantically smooth while to 
  expand it with an ironic evaluation when necessary, as shown in sixth sentence.
  
  In this paper, we present a textual adversarial attacking method orienting the NLP task of sentiment analysis by 
  rewriting a straightforward sentence into an ironic sentence, namely IAE (Irony-based Adversarial Examples). To the best of our knowledge, 
  we are the first to use irony for textual adversarial examples generation. We summarize our major contributions as follows:
  \begin{itemize}
	\item We propose IAE, a strategy based on the concept of a rhetorical device called irony for generating textual adversarial examples, which does not need to prepare irony corpus.
	\item We show that the performance of various deep learning models substantially drops for sentiment analysis tasks when attacked by IAE.
	\item We show that humans are only mildly or not at all affected by irony in contrast to NLP systems.
  \end{itemize}

 \section{Literature review}
 Our work connects to two strands of literature: textual adversarial examples and irony generation. 
 
\noindent\textbf{Textual Adversarial Examples}
Existing textual adversarial attack models can be categorized into character-level, word-level, and sentence-level according to the perturbation levels of their adversarial examples. 
 
Character-level attacks disrupt the process of converting natural language text into numerical representations that computers can process, thereby causing model decision shifts. 
The manifestation of character-level attacks varies across different linguistic environments. 
In English, character-level attacks often exploit visual perturbations, such as inserting\cite{formento-etal-2023-using}, deleting, swapping, and modifying\cite{Eger19} letters within words to create artificially constructed spelling errors. 
In the Chinese context, handwriting errors on paper do not occur in electronic input based on input methods. 
Therefore, character-level attacks in the Chinese environment often manifest as the use of homophones for substitution\cite{homonym,DBLP:journals/access/ChengCGPZ20} or visual decomposition of characters\cite{DBLP:journals/kais/OuYTC22}.

Word-level adversarial attacks achieve a shift in the semantic vector of the sample by perturbing the input sample at the word level, causing it to cross the decision boundary and thus leading to incorrect model outputs. 
Word substitution, as the core method of this strategy, includes various word replacement means such as word vector similarity\cite{DBLP:conf/aaai/JinJZS20}, synonyms\cite{RenDHC19}, and language model scoring\cite{zhang19}. 
Word-level adversarial attacks do not break the grammatical rules of the text and retain the original semantics to the greatest extent, thus performing better in terms of adversarial text quality and attack success rate. Coupled with the use of language models for control, it also ensures the fluency and smoothness of adversarial texts. 
Among them, text attacks based on synonym substitution have strong semantic retention and grammatical coherence, belonging to the most threatening category of text adversarial attacks, which have attracted widespread attention from researchers. 

Sentence-level adversarial attacks treat the entire original input sentence as the object of perturbation, carefully reconstructing the text content, that is, generating adversarial text that has the same semantics as the original input but causes the victim model to make decision errors. 
Common sentence-level adversarial attack methods include encoding and then re-decoding\cite{DBLP:conf/emnlp/HanZJT20}, adding irrelevant sentences\cite{DBLP:conf/ijcai/0002LSBLS18}, paraphrasing\cite{xu-etal-2021-grey}, etc. 

 \noindent\textbf{Irony Generation}
The field of irony generation, particularly within the Chinese linguistic context, remains largely unexplored, with limited research and development dedicated to this area. Zhu et al.~\cite{abs-1909-06200} proposed a novel method that integrates reinforcement learning with style transfer techniques to generate ironic text. Their approach relies on a carefully designed reward system to guide the model towards producing text that effectively conveys irony. This method demonstrates the potential of combining advanced machine learning techniques with stylistic adjustments to achieve the nuanced expression of irony. Veale et al.~\cite{Veale18} took a different route by exploring knowledge-based systems and shallow linguistic techniques, which they term "mere re-generation," for irony generation. This approach leverages existing knowledge structures and simple linguistic manipulations to introduce ironic elements into the text. While this method may not delve deeply into the complexities of language, it offers a more straightforward and potentially more accessible avenue for irony generation. In the closely related domain of sarcasm, Mishra et al.~\cite{Mishra19} presented a framework that utilizes reinforced neural sequence-to-sequence learning coupled with information retrieval strategies for sarcasm generation. 
 
To the best of our knowledge, our work represents the first instance of leveraging irony for the generation of textual adversarial examples. This application of irony in adversarial machine learning is groundbreaking, as it introduces a new dimension to the field of natural language processing security. It serves as a testament to the importance of understanding and incorporating advanced linguistic features, such as irony, into machine learning models to enhance their resilience against adversarial attacks. 

\section{Problem statement}
We assume access to a corpus of labeled sentences $D=\{(s_{1},p_{1}),...,(s_{n},p_{n})\}$, where $s_{i}$ is a sentence and 
$p_{i} \in L$, the set of possible emotional polarity, i.e., $L$ = \{positive, negative\}. 
We define $s^{p}=(c, e, d)$, a sentence with emotional polarity $p$, where $c$ is the central word of the sentence, 
$e$ is the evaluation word that evaluating the central word $c$, and $d$ is the detailed description of the evaluation. 
On this basis, we define emotional sentence $s^{p}$ as a straightforward sentence or an ironic sentence while the evaluation $e$ 
have emotional polarity $p'$, while collocating with $c$, and $p = p'$, or $p \neq p'$.

Generally, the irony is a negative sentence exhibiting positive evaluation. Thus, our goal is to build a model that takes as 
input sentence $s$, a negative emotional sentence exhibiting negative evaluation $e^{\textrm{neg}}$, and outputs a sentence $s'$ 
that retains the negative emotional polarity while exhibiting positive evaluation $e^{\textrm{pos}}$.
Note that the concept of evaluation word we use is not equivalent to the sentiment word while sentiment word is an adjective with a clear emotional polarity. 
The emotional polarity of an evaluation word should be determined by the central word with which the evaluation word collocates.

\section{Approach}
\begin{figure*}[ht]
  \centering
  \includegraphics[width=0.7\textwidth]{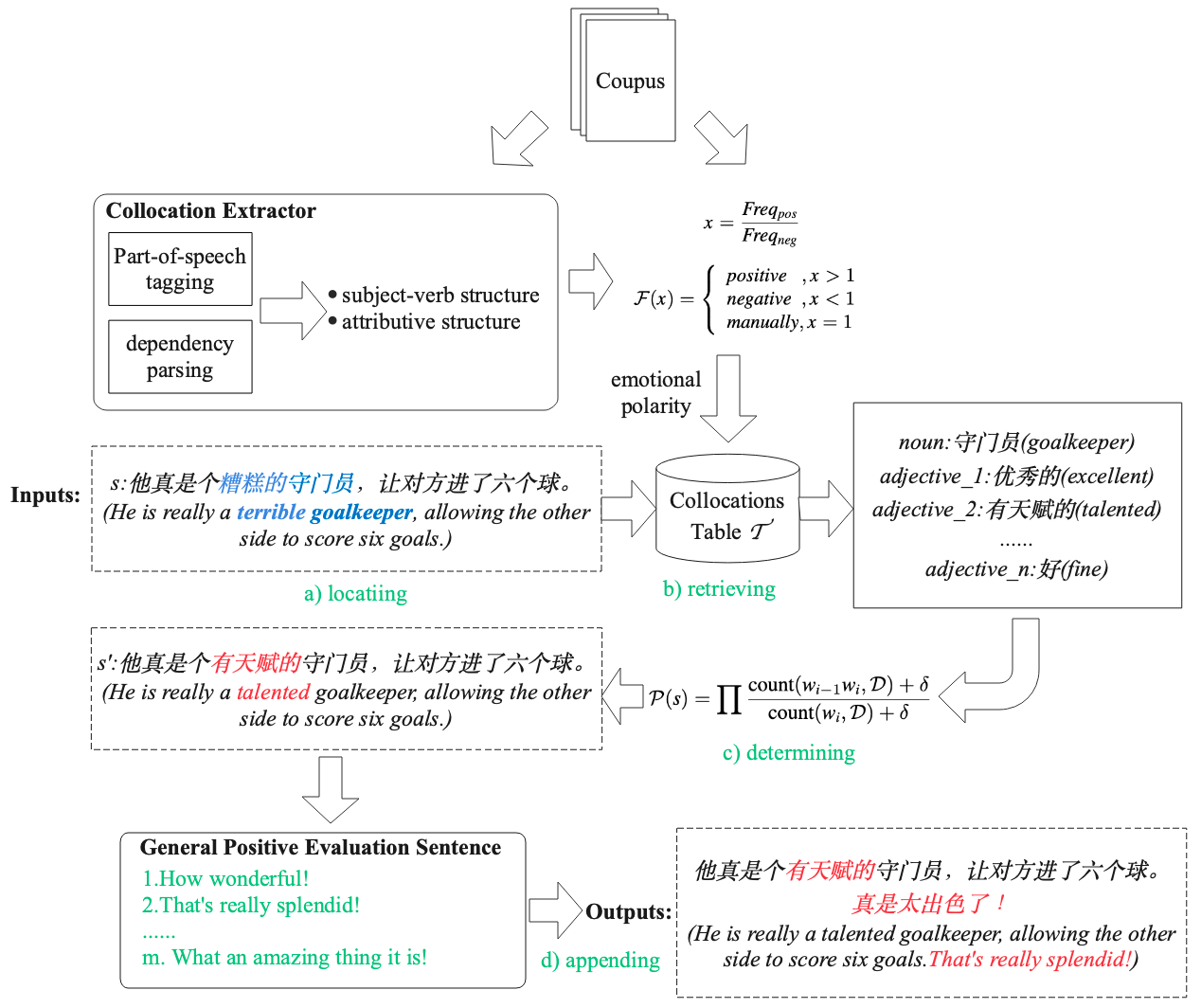}
  \caption{\label{fig:flow chart} An overview of our proposed IAE generator. }
  \end{figure*}

In this section, we detail our irony-based textual adversarial attacking method, incorporating three parts:
1) an extractor of collocations between nouns and adjectives, 
2) a strategy for evaluation word substitution, and 
3) a strategy for ironic evaluation sentence generation.
An overview of our IAE generator is shown in Fig.~\ref{fig:flow chart}. 
Generally, it takes straightforward text as inputs and outputs ironic text.
First, the central word and relevant evaluation word will be located, and then the evaluation word will be substituted with an opposite evaluation word among all possible alternatives,
Finally, an appropriate ironic evaluation sentence, determined by local model, will be appended to the text for strengthening the effect of irony.

Next, we describe the details of each component of IAE generator.

\subsection{Collocation extractor}
\label{sec:extractor}

We design a collocations extractor to establish noun-adjective collocations tables,
which also reveals probable emotional polarity between a noun with all collocated adjectives, 
as a library of alternatives for evaluation word substitution (see section~\ref{sec:substitution}). 

A host of observations were made on Chinese corpus with part-of-speech tagging and dependency parsing, and we found the
noun-adjective collocations in a Chinese sentence are supposed to form the following two kinds of dependencies:
1) a subject-verb structure, or
2) an attributive structure (see examples in Table~\ref{table:collocations}).
Note that the results of dependency parsing in Chinese may be different from English due to the differences in the two kinds of syntax rules. 
e.g., the words ``\emph{weather}'' and ``\emph{good}'' are supposed to form a subject-predicative in English instead of a subject-verb.

Then we can extract plenty of collocations from a large corpus through the observations above, but the next key question is how to determine the emotional polarity of each noun-adjective collocation.
Although we can use advanced sentiment analysis models to determine the overall emotional polarity of the sentence from which a noun-adjective collocation extracted,
it is no guarantee the emotional polarity of a noun-adjective collocation will be consistent with the whole sentence. 
But, intuitively, the emotional polarity of a collocation should probably be positive if it mostly appears in sentences with a positive overall emotional polarity rather than negative.
Hence, the polarity of collocation can be inferred by the following formulas:

\begin{equation}
	x =\frac{\textrm{Freq}_{\textrm{pos}}}{\textrm{Freq}_{\textrm{neg}}}
   \end{equation}
   
   \begin{equation}
	 F(x)=
	 \left\{ 
	 \begin{array}{l}
	   \textrm{positive} \ \ \ , x>1 \\
	   \textrm{negative} \ \ , x<1 \\
	   \textrm{manually}, x=1 \\
	 \end{array}
	 \right.
	 \\
	 \end{equation}
   
   Where $\textrm{Freq}_{\textrm{pos}}$ and $\textrm{Freq}_{\textrm{neg}}$ are the frequencies of a collocation appearing in sentences with an emotional polarity of positive or negative respectively.
   The emotional polarity of a collocation is supposed to be positive when $x>1$, or negative when $x<1$, or decided manually when the result of $x$ 
   happens to be $1$.
   
   Therefore, the noun-adjective collocations table, denoted as $T$, can be established by collecting collocations by 
   dependency parsing and inferring their emotional polarities by counting and comparing the numbers of each emotional polarity of 
   the sentences in which they occur.

 \subsection{Evaluation word substitution}
 \label{sec:substitution}
 The strategy for evaluation word substitution is the most important procedure to convert a straightforward sentence $s$ to an ironic sentence $s'$
 while $s'$ has the evaluation word $e$ with emotional polarity $p'$, which is opposite to the emotional polarity $p$ of
 the whole sentence. Next, we describe our evaluation word substitution step by step.
 
 \noindent\textbf{Locating.} \ At the very beginning, the pairs of central word and relevant evaluation word are located 
 by using part-of-speech tagging and dependency parsing together, which is similar to the strategy of extracting noun-adjective 
 collocation (see secttion~\ref{sec:extractor}). 
 
 \noindent\textbf{Retrieving.} \ The alternatives are retrieved among the table $T$ by using central word $c$ 
 as an index. The whole procedure will terminate and return a general evaluation word (e.g., \begin{CJK}{UTF8}{gbsn}``不错''\end{CJK}, which is analog to ``fine'' in English) 
 as the result while the central word does not exist or none of the positive evaluation words are retrieved.
 
 \noindent\textbf{Determining.} \ To determine what alternative evaluation word to substitute original, our strategy is to evaluate the 
 quality (i.e., probability of sentence) of all alternative sentences $S'$ by N-gram language model while combining any possible collocation of central word and alternative evaluation word. 
 Formally, for any $s' \in S'$, the probability is calculated by the following formula:
 \begin{equation}
 P(s)=\prod\frac{\textrm{count}(w_{i-1}w_{i},D)+\delta}{\textrm{count}(w_{i},D)+\delta}
 \end{equation}
 where $w_{i}$ is the $i$-th word in $s'$, $w_{i-1}w_{i}$ is the sequence composed of $w_{i-1}$ and $w_{i}$ sequentially, 
 $\textrm{count}(.,D)$ denotes the numbers of times a word or a sequence appears in $D$, 
 and $\delta$ is an additive smoothing parameter for the situation that some words are just not appearing in $D$. 
 In practice, the smoothing parameter $\delta$ can be set to 1 empirically. The alternative sentence $s'$ with the highest probability among $S'$ 
 will be determined as the result of evaluation word substitution.

 \begin{table*}[]
	\centering
	\caption{\label{table:collocations} Examples of sentences containing noun-adjective collocations and dependencies.}
	\begin{tabular}{lrl}
	\toprule
	\makecell[c]{\textbf{Sentences}} & \makecell[c]{\textbf{Collocations}} & \makecell[c]{\textbf{Dependencies}} \\  
	\midrule
	\specialrule{0em}{1pt}{1pt}
	\makecell[l]{
	  \begin{CJK}{UTF8}{gbsn}\textcolor{blue}{天气}这么\textcolor{blue}{好}，应该出去透透空气。\end{CJK}\\
	  The \textcolor{blue}{weather} is so \textcolor{blue}{good} for enjoying fresh air.
	}&
	\makecell[c]{\begin{CJK}{UTF8}{gbsn}天气，好\end{CJK}\\
	weather, good
	}& 
	\makecell[c]{subject-verb}\\  
	\midrule
	\specialrule{0em}{1pt}{1pt}
	\makecell[l]{
	\begin{CJK}{UTF8}{gbsn}那个\textcolor{blue}{男人}真\textcolor{blue}{帅}。\end{CJK}\\
	That \textcolor{blue}{man} is so \textcolor{blue}{handsome}. 
	}& 
	\makecell[c]{
	  \begin{CJK}{UTF8}{gbsn}男人，帅\end{CJK}\\
	  man, handsome
	}& 
	\makecell[c]{subject-verb}\\  
	\midrule
	\specialrule{0em}{1pt}{1pt}
	\makecell[l]{
	  \begin{CJK}{UTF8}{gbsn}\textcolor{blue}{优美}的\textcolor{blue}{音乐}可以给人们带来享受。\end{CJK}\\
	  \textcolor{blue}{Beautiful} \textcolor{blue}{music} can bring people enjoyment.
	}& 
	\makecell[c]{
	  \begin{CJK}{UTF8}{gbsn}音乐，优美\end{CJK}\\
	  music, beautiful
	  }& 
	\makecell[c]{attributive}\\  
	\midrule
	\specialrule{0em}{1pt}{1pt}
	\makecell[l]{
	  \begin{CJK}{UTF8}{gbsn}她招待我们吃了一顿\textcolor{blue}{可口}的\textcolor{blue}{午餐}。\end{CJK}\\
	  She served us a \textcolor{blue}{delicious} \textcolor{blue}{lunch}.
	}& 
	\makecell[c]{
	  \begin{CJK}{UTF8}{gbsn}午餐，可口\end{CJK}\\
	  lunch, delicious
	}& 
	\makecell[c]{attributive}\\  
	\bottomrule
	\end{tabular}
	\end{table*}
 
\subsection{Ironic evaluation appending}
 \label{sec:appending}
 Reversing the result of sentiment analysis by substituting the evaluation alone is often difficult while the context 
 still exhibits original emotional polarity. But this problem can be solved by appending an evaluation, which is opposite to 
 the polarity of real emotion for strengthening the ironic effect. 
 
 It is easy to construct positive evaluations by composing positive adjectives and other grammatical constituents according to sentence patterns. 
 However, the problems are how to choose an evaluation and how to guarantee the semantic smoothness of the whole sentence after evaluation appending. 
 
 Our strategy is to construct general positive evaluations, which can collocate with almost objects and guarantee the semantic smoothness, 
 as much as possible, and then to determine an evaluation appending to $s'$.
 
 Inspired by the substitute black box attack (SBA)~\cite{SBA} which is utilizing the transferability of adversarial examples, 
 we consider training the local model to substitute the victim model, then testing each alternative on the local model, and 
 finally selecting the evaluation while the local model outputs a wrong prediction after appending.
 For the case that there is no effective adversarial example on the local model, we consider choosing the longest one.
 
 After determining the ironic evaluation, which is supposed to append to the sentence $s'$, the final IAE is generated completely. 

 \section{Experiments and Results}
 In this section, we conduct comprehensive experiments to evaluate our IAE on the tasks of sentiment analysis. 

 \subsection{Datasets and Victim Models}
 We evaluate our IAE on the public reviews of Meituan\footnote{Meituan is a platform for ordering takeaway, which contains positive and negative user reviews} 
 and Amazon. The five-star and one-star reviews are taken as positive and negative text respectively. 
 Because our IAE is only applicable to the examples with negative emotional polarity, we randomly select 500 examples with negative emotional polarity from each dataset as the test set, 
 and then we divide the remaining examples into two balanced parts for training local and victim models.
 Details of the datasets are shown in Table~\ref{table:dataset}, where ``Class \#'' refers to the number of labels,
 ``Max. \#W'' means maximum length of sentences (number of words), ``Min. \#W'' means minimum length of sentences (number of words),
 and ``Avg. \#W'' means average length of sentences (number of words), ``P. \#'' and ``N. \#'' signify the number of text exhibiting positive and negative emotional polarity respectively.
 
 Besides, for comprehensive noun-adjective collocations extracting (see section~\ref{sec:extractor}), 
 we collected 30111 nouns and 114383 related collocations from serveral Chinese corpus, including reviews on Meituan and Amazon, Sina weibo\footnote{a twitter like Chinese online platform} comments, and online News corpus. 
 For each noun, there are 1115 collocations at most and 1 collocation at least, with an average of 3.7.
 
 We choose three popular models for text classification, namely TextCNN~\cite{TextCNN}, Bidirectional LSTM (BiLSTM)~\cite{BidLSTM} 
 and a fine-tuned BERT~\cite{BERT}, used for evaluating our IAE. 
 TextCNN has three convolutional filters of different kernel sizes (3, 4, 5), and their outputs are concatenated, pooled and fed to a fully-connected layer followed by an output layer.
 BiLSTM is composed of a 128-dimenional bidirectional LSTM layer, a dropout layer using a drop rate of 0.5, and an output layer.
 BERT is obtained by fine-tuning the Chinese BERT-Base model with 12-layer, 768-hidden, and 12-heads released by Google.
 The optimizer, learning rate, and loss function of all models are set to adam, 0.01, and cross-entropy respectively.
 Besides, we implement Chinese word segmentation, part of speech tagging, and dependency parsing using the third-party library released by Harbin Institute of technology~\cite{DBLP:journals/corr/abs-2009-11616}.

 \subsection{Baseline methods}
 We implement two baseline methods based on important word substitution and compared them with ours for proving the contribution of this work.
 The two baseline methods are 1) visual-based substitution~\cite{Eger19}, which means substitute important words with visual similar chart, 
 and 2) homonym-based substitution~\cite{homonym}, which means substitute important words with others pronounced the same way but have different meanings.
 The important words refer to the words in the input text that make the most contribution to the model decision and the calculation algorithm of important words adopts~\cite{DBLP:conf/ndss/LiJDLW19}.  
 \begin{table*}[htbp]
    \centering
    \caption{\label{table:dataset} Statistics for the datasets.}
    \begin{tabular}{ccccccc}
    \toprule
    \makecell[c]{\textbf{Dataset}}&\makecell[c]{\textbf{Class \#}}&\makecell[c]{\textbf{Max. \#W}}&\makecell[c]{\textbf{Min. \#W}}&\makecell[c]{\textbf{Avg. \#W}}&\makecell[c]{\textbf{P. \#}}&\makecell[c]{\textbf{N. \#}}\\
    \midrule
    \makecell[c]{Meituan}&\makecell[c]{2}&\makecell[c]{237}&\makecell[c]{2}&\makecell[c]{18.96}&\makecell[c]{6000}&\makecell[c]{6500}\\
    \makecell[c]{Amazon}&\makecell[c]{2}&\makecell[c]{858}&\makecell[c]{2}&\makecell[c]{23.71}&\makecell[c]{6000}&\makecell[c]{6500}\\
    \bottomrule
    \end{tabular}
    \end{table*}

    \begin{table*}[htbp]
   \centering
   \caption{\label{table:attacking performance} Performance of victim models under attacking of IAE and two baseline methods on Meituan review dataset. }
   \begin{tabular}{|c|c|c|ccc|c|}
     \hline
     \multirow{2}{*}{\textbf{Dataset}} &
       \multirow{2}{*}{\textbf{Method}} &
       \multirow{2}{*}{\textbf{Local Model}} &
       \multicolumn{3}{c|}{\textbf{Victim model}} &
       \multirow{2}{*}{\textbf{WMD}} \\ \cline{4-6}
                              &                                &         & \multicolumn{1}{c|}{TextCNN} & \multicolumn{1}{c|}{BidLSTM} & Bert  &     \\ \hline
     \multirow{10}{*}{Meituan} & Origin                         & N/A     & \multicolumn{1}{c|}{0.885}   & \multicolumn{1}{c|}{0.894}   & 0.934 & N/A \\ \cline{2-7} 
                              & \multirow{3}{*}{Visual-based}  & TextCNN & \multicolumn{1}{c|}{0.312}   & \multicolumn{1}{c|}{0.702}   & 0.880 & 1.497    \\ \cline{3-7} 
                              &                                & BidLSTM & \multicolumn{1}{c|}{\textbf{0.260}}   & \multicolumn{1}{c|}{0.670}   & 0.892 & 1.787    \\ \cline{3-7} 
                              &                                & Bert    & \multicolumn{1}{c|}{0.610}   & \multicolumn{1}{c|}{0.818}   & 0.860 & 0.384    \\ \cline{2-7} 
                              & \multirow{3}{*}{Homonym-based} & TextCNN & \multicolumn{1}{c|}{0.326}   & \multicolumn{1}{c|}{0.694}   & 0.854 & 1.551    \\ \cline{3-7} 
                              &                                & BidLSTM & \multicolumn{1}{c|}{0.304}   & \multicolumn{1}{c|}{0.698}   & 0.856 & 1.784    \\ \cline{3-7} 
                              &                                & Bert    & \multicolumn{1}{c|}{0.606}   & \multicolumn{1}{c|}{0.832}   & 0.846 & 0.411    \\ \cline{2-7} 
                              & \multirow{3}{*}{Ours}          & TextCNN & \multicolumn{1}{c|}{0.464}   & \multicolumn{1}{c|}{\textbf{0.204}}   & 0.456 & 1.197    \\ \cline{3-7} 
                              &                                & BidLSTM & \multicolumn{1}{c|}{0.324}   & \multicolumn{1}{c|}{0.416}   & 0.542 & 1.041    \\ \cline{3-7} 
                              &                                & Bert    & \multicolumn{1}{c|}{0.700}   & \multicolumn{1}{c|}{0.876}   & \textbf{0.370} & 0.808    \\ \hline
     \multirow{10}{*}{Amazon} & Origin                         & N/A     & \multicolumn{1}{c|}{0.900}   & \multicolumn{1}{c|}{0.936}   & 0.944 & N/A \\ \cline{2-7} 
                              & \multirow{3}{*}{Visual-based}  & TextCNN & \multicolumn{1}{c|}{0.564}   & \multicolumn{1}{c|}{0.664}   & 0.870 & 2.570 \\ \cline{3-7} 
                              &                                & BidLSTM & \multicolumn{1}{c|}{0.346}   & \multicolumn{1}{c|}{0.400}   & 0.914 & 2.643 \\ \cline{3-7} 
                              &                                & Bert    & \multicolumn{1}{c|}{0.860}   & \multicolumn{1}{c|}{0.830}   & 0.836 & 2.480 \\ \cline{2-7} 
                              & \multirow{3}{*}{Homonym-based} & TextCNN & \multicolumn{1}{c|}{0.648}   & \multicolumn{1}{c|}{0.702}   & 0.844 & 2.559 \\ \cline{3-7} 
                              &                                & BidLSTM & \multicolumn{1}{c|}{0.544}   & \multicolumn{1}{c|}{0.550}   & 0.870 & 2.605 \\ \cline{3-7} 
                              &                                & Bert    & \multicolumn{1}{c|}{0.842}   & \multicolumn{1}{c|}{0.826}   & 0.834 & 2.472 \\ \cline{2-7} 
                              & \multirow{3}{*}{Ours}          & TextCNN & \multicolumn{1}{c|}{\textbf{0.338}}   & \multicolumn{1}{c|}{\textbf{0.322}}   & 0.602 & 2.344 \\ \cline{3-7} 
                              &                                & BidLSTM & \multicolumn{1}{c|}{0.720}   & \multicolumn{1}{c|}{0.728}   & \textbf{0.538} & 2.341\\ \cline{3-7} 
                              &                                & Bert    & \multicolumn{1}{c|}{0.880}   & \multicolumn{1}{c|}{0.886}   & 0.670 & 2.379 \\ \hline
     \end{tabular}
   \end{table*}

\begin{table}[htbp]
   \centering
   \caption{\label{table:quality} Human evaluation of emotional correctness and grammar smoothness.}
   \begin{tabular}{|c|cc|cc|}
     \hline
     \multirow{2}{*}{\textbf{Dataset}}&\multicolumn{2}{c|}{\textbf{Emotional Correctness}}&\multicolumn{2}{c|}{\textbf{Grammar Smoothness}}\\
     \cline{2-5}
     \makecell[c]{}&\makecell[c]{Origin}&\makecell[c]{IAE}&\makecell[c]{Origin}&\makecell[c]{IAE}\\
     \hline
     \makecell[c]{Meituan}&\makecell[c]{91}&\makecell[c]{86}&\makecell[c]{4.25}&\makecell[c]{3.50}\\
     \makecell[c]{Amazon}&\makecell[c]{90}&\makecell[c]{84}&\makecell[c]{4.50}&\makecell[c]{3.75}\\
     \hline
   \end{tabular}
 \end{table}
 
\subsection{Attack performance}
 The attack performance results of our IAE and two baseline methods are shown in Table~\ref{table:attacking performance}. 
 Note that only examples labeled with negative are used for test as the adversarial attack based on irony is only applicable to the examples with negative emotional polarity.
 We observe the adversarial examples generated by our irony-based attack cause the victim models to fail more seriously than the baseline methods in most conditions in most conditions.
 Specifically, the visual-based and homonym-based attack can hardly fool Bert models while our method can cause the accuracy of Bert from 89.8\% to 37.0\% at most,
 besides, the Word Mover’s Distances~\cite{WMD} between our IAE and clean examples are always smaller than those between adversarial examples generated by baseline methods and clean examples.
 
 \subsection{Human evaluation}
 We ask 4 students with native Chinese language skill to evaluate the emotional correctness and semantic smoothness of successful IAE generated from Meituan and Amazon reviews. 
 Specifically, we randomly select 100 IAE and 100 clean examples, and every student needs to evaluate the mixture of them.
 
 For evaluating emotional correctness, each student evaluates the true emotional polarity of each example and it is annotated as positive (negative) if two or more students evaluate a example as positive (negative).
 An extra human evaluator would participate in the evaluation if there are equal numbers of different evaluation on emotional polarity.
 
 For evaluating semantic smoothness, each student scores the semantic smoothness of each example with Likert scale ranging from 1 to 5 
 while 1 and 5 mean the semantics of a example is completely confused or fluent separately.
 We summarized the evaluation of all students and averaged the semantic smoothness of the IAE and the clean examples respectively. 
 
 The results are shown in Table~\ref{table:quality} and it shows that a lightly lower emotional correctness and semantic smoothness in IAE than clean examples,
 but the emotional correctness and semantic smoothness of IAE still reach 86 and 3.75 respectively.
 
 \section{Discussion}
 We studied how to regard irony as a textual adversarial perturbation in Chinese and it proved effective in sentiment analysis. 
 There are differences between Chinese and other languages in grammar and habits, however, irony, as a rhetorical device in almost all languages, 
 could be utilized as a general way of textual adversarial perturbation.
 
 The experiment of training the local model for generating effective adversarial examples also reveals some properties of transferability.
 First, the transfers between two models are non-symmetric. As we can see, the accuracy of victim model BERT is $54.2\%$ when generated IAE from local model BidLSTM, 
 however, the accuracy of victim model BidLSTM is $87.6\%$ when generated IAE from local model BERT while testing on Meituan reviews dataset. 
 It is similar to the findings in the study of the transferability of image adversarial examples~\cite{DBLP:conf/acml/WuZ20}, even though we focus on the text field.
 Second, the adversarial examples generated from the high-accuracy models may be less transferable. As we can see, BERT is the most accurate model among all models we use, however, 
 the adversarial examples generated from BERT hardly mislead other models.
 
 We also found there are three major types of weaknesses in our methods, which affect the attacking performances. 
 For analysis of the weaknesses, we sampled 100 failed IAE which mislead the victim model unsuccessfully or lose original sentiment. 
 We found that 26\% of the failures are due to the long length of input text which is more than 50 Chinese characters, 
 38\% of the failures are due to the weak correlation between evaluative sentence and context, 
 29\% of the failures are due to the imperfection of part-of-speech tagging and dependency parsing tools, 
 and the remaining 7\% of the failures have no significant type. 
 
 The first type of failure is due to the obvious fact that the longer the text, the more negative content it contains, 
 so it is difficult to change the label of model prediction by substituting an evaluation word or appending a generally positive evaluation sentence. 
 
 The second type of failure is due to the weak correlation between the evaluative sentence and the context description. 
 For example, for the sentence ``\begin{CJK}{UTF8}{gbsn}菜真的很难吃，还是去其他店吃好些\end{CJK}'' (The food is really unpalatable, and it's better to go to another restaurant), 
 where the context is not a correlational detail description to the evaluation of food, it is inappropriate to substitute the negative evaluation word ``\begin{CJK}{UTF8}{gbsn}难吃\end{CJK}'' (unpalatable) 
 to a positive word ``\begin{CJK}{UTF8}{gbsn}好吃\end{CJK}'' (delicious) otherwise the emotional polarity of the text will change completely.
 
 The third type of failure is due to the dependency analysis tools, which is unable to analyze the dependency correctly all the time, 
 while it is necessary to locate the evaluation word with part-of-speech tagging and dependency parsing.
 
 \section{Conclusion and future work}
 In this paper, we have introduced Irony-based Adversarial Examples (IAE), a novel method for generating adversarial text by transforming straightforward sentences into ironic ones. 
 Our research has made several significant contributions to the field of adversarial attack. 
 Firstly, we have introduced IAE as a strategy for generating textual adversarial examples that leverages irony. 
 This method is innovative in that it does not depend on pre-existing irony corpora, thereby offering a flexible instrument for creating adversarial text across a spectrum of NLP tasks. 
 Secondly, we have demonstrated empirically that the performance of several deep learning models on sentiment analysis tasks is markedly compromised when confronted with IAE attacks. 
 This result highlights the vulnerability of current NLP systems to adversarial manipulations facilitated through irony. Thirdly, we have compared the effects of IAE on human judgment versus NLP systems, revealing a notable difference in susceptibility. 
 Our findings indicate that humans are relatively more resilient to the influence of irony in text, contrasting with the performance of NLP models.
 
 Our future work will focus on enhancing the performance of IAE in longer texts and improving its generalization capabilities across different languages. 
 This will involve addressing the complexities associated with maintaining ironic integrity over extended passages and adapting to the nuances of various linguistic contexts. 
 Additionally, we are intrigued by the prospect of integrating more rhetorical devices into textual adversarial perturbations, beyond irony. 
 For instance, exploring the use of metaphors to disrupt machine reading comprehension presents an exciting avenue for further research. 
 By expanding the repertoire of rhetorical strategies employed in adversarial text generation, we aim to deepen our understanding of the interplay between language, context, and machine learning models, ultimately contributing to the development of more robust and nuanced NLP systems.

\bibliographystyle{IEEEtran} 
\bibliography{bibliography}


\begin{IEEEbiography}[{\includegraphics[width=1in,height=1.25in,clip,keepaspectratio]{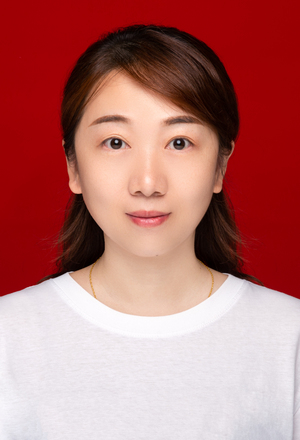}}]{Xiaoyin Yi} received the M.S. degree in computer science from the Chongqing University of Posts and Telecommunications, China. 
    She is currently a teacher with Chongqing College of Mobile Communication and pursuing the Ph.D. degree with Chongqing University of Posts and Telecommunications, China. 
    Her research interests include
    cybersecurity within AI and AI security.
  \end{IEEEbiography}

  \begin{IEEEbiography}[{\includegraphics[width=1in,height=1.25in,clip,keepaspectratio]{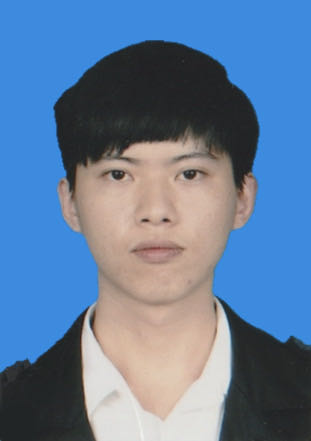}}]{Jiacheng Huang} received the M.S. degree in instructional technology from the Hubei Normal University, China. 
    He is currently pursuing the Ph.D. degree with Chongqing University of Posts and Telecommunications, China. 
    His research interests include
    cybersecurity within AI, natural language processing, and AI security.
  \end{IEEEbiography}

\EOD

\end{document}